\pdfoutput=1

\documentclass[11pt]{article}

\usepackage[final]{acl}

\usepackage{times}
\usepackage{latexsym}

\usepackage[T1]{fontenc}

\usepackage[utf8]{inputenc}

\usepackage{microtype}

\usepackage{epsfig}
\usepackage{amsmath}
\usepackage{amssymb}
\usepackage{microtype}
\usepackage{xspace}
\usepackage{subfig}
\usepackage{graphicx,wrapfig}
\usepackage{booktabs}
\usepackage{makecell}
\usepackage{framed}
\usepackage{multirow}
\usepackage[linesnumbered,algoruled,boxed,noend]{algorithm2e}
\usepackage{enumitem}

\usepackage{pifont}
\newcommand{\method}{\textsc{FewVLM}}%

\newcolumntype{L}[1]{>{\raggedright\let\newline\\\arraybackslash\hspace{0pt}}m{#1}}
\newcolumntype{C}[1]{>{\centering\let\newline\\\arraybackslash\hspace{0pt}}m{#1}}
\newcolumntype{R}[1]{>{\raggedleft\let\newline\\\arraybackslash\hspace{0pt}}m{#1}}

\SetKwInput{KwInput}{Input}
\SetKwInput{KwOutput}{Output}
\SetEndCharOfAlgoLine{}

\newcommand{\para}[1]{\smallskip\noindent\textbf{#1}}

\title{\textit{A Good Prompt Is Worth Millions of Parameters:} \\Low-resource Prompt-based Learning for Vision-Language Models}

\author{
    Woojeong Jin$^1$\thanks{\xspace\xspace Work was mainly done while interning at Microsoft Azure AI.} \quad Yu Cheng$^2$  \quad Yelong Shen$^2$ \quad Weizhu Chen$^2$ \quad Xiang Ren$^1$\\
    $^1$University of Southern California \quad $^2$Microsoft Corporation \\
    {\small \texttt{\{woojeong.jin,xiangren\}@usc.edu} \texttt{\{yu.cheng,yelong.shen,wzchen\}@microsoft.com}}\\
}

\begin{document}
\maketitle
\begin{abstract}

Large pre-trained vision-language (VL) models can learn a new task with a handful of examples and generalize to a new task without fine-tuning.
However, these VL models are hard to deploy for real-world applications due to their impractically huge sizes and slow inference speed.
To solve this limitation, we study prompt-based low-resource learning of VL tasks with our proposed method, \method, relatively smaller than recent few-shot learners.
For \method, we pre-train a sequence-to-sequence transformer model with prefix language modeling (PrefixLM) and masked language modeling (MaskedLM).
Furthermore, we analyze the effect of diverse prompts for few-shot tasks.
Experimental results on VQA show that \method\xspace with prompt-based learning outperforms Frozen \cite{tsimpoukelli2021multimodal} which is 31$\times$ larger than \method\xspace by 18.2\% point and achieves comparable results to a 246$\times$ larger model, PICa~\cite{yang2021empirical}.
In our analysis, we observe that (1) prompts significantly affect zero-shot performance but marginally affect few-shot performance, (2) models with noisy prompts learn as quickly as hand-crafted prompts given larger training data, and (3) MaskedLM helps VQA tasks while PrefixLM boosts captioning performance. Our code is publicly available at \url{https://github.com/woojeongjin/FewVLM}



\end{abstract}

\section{Introduction}

\begin{figure}[tb!]
    \centering
    {\includegraphics[width=0.9\columnwidth]{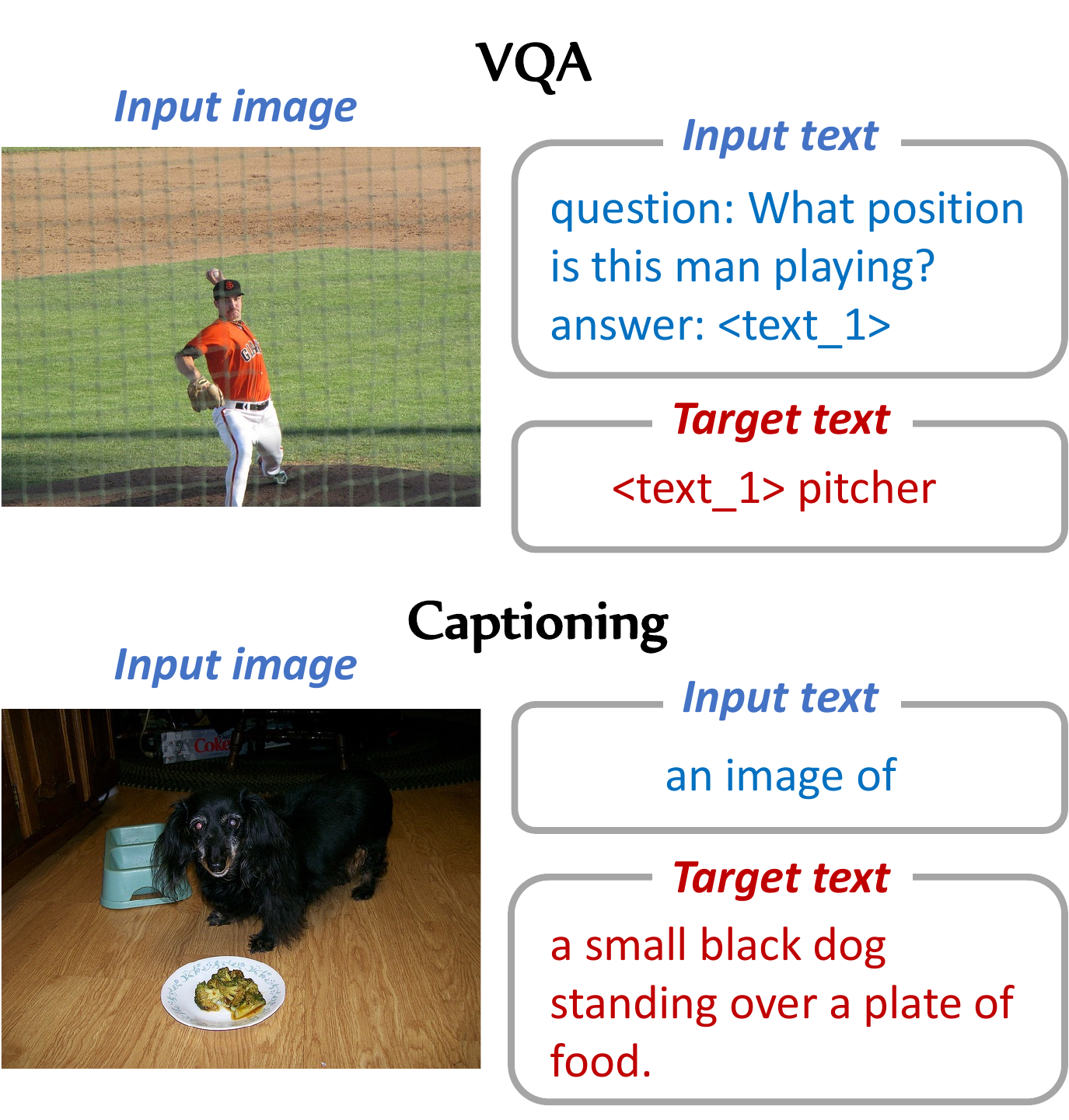}}
    \caption{\textbf{Examples of VQA and Captioning tasks.} In our setup, we convert the tasks into generative tasks in which models need to generate target text given input text and an image. 
    }
    \label{fig:fewshot}
\end{figure}

\begin{figure*}[tb!]
    \centering
    {\includegraphics[width=0.8\linewidth]{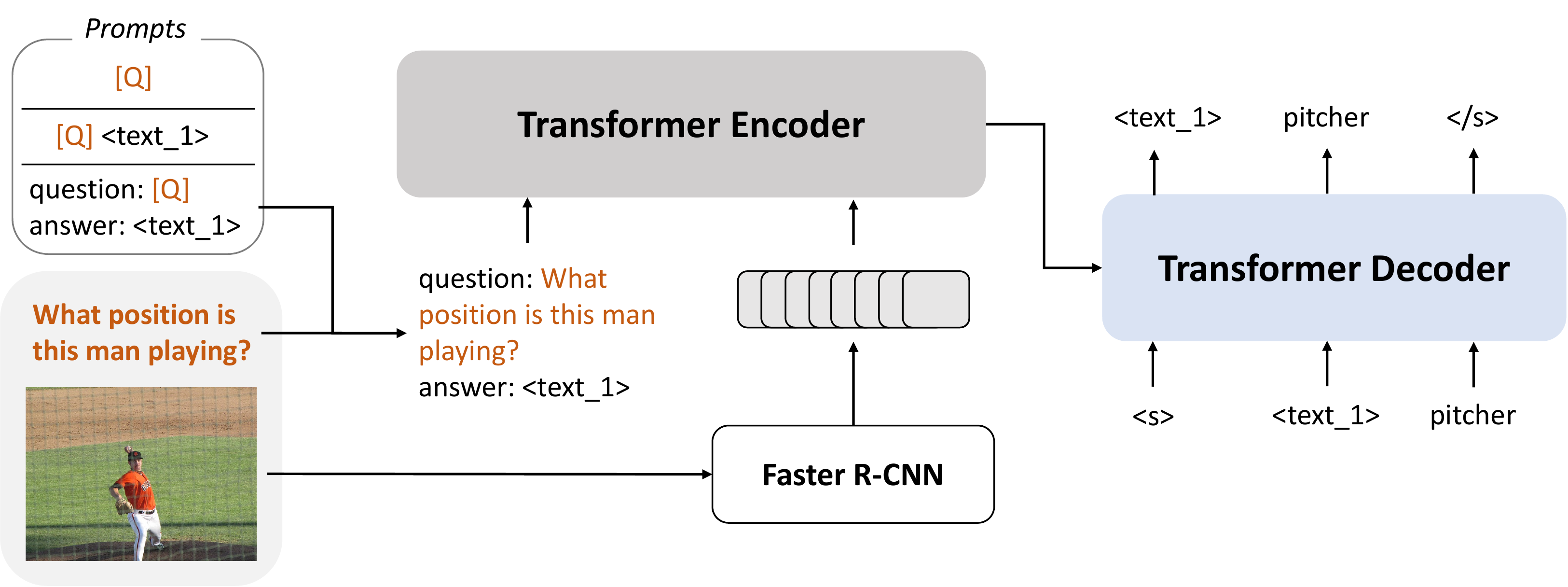}}
    \caption{\textbf{Illustration of \method.} This shows inference of \method\xspace with prompt-based learning. Given a prompt template, we convert the question text into input text. The prompt helps the model generate correct answers.
    }
    \label{fig:illust}
\end{figure*}

Fine-tuning large pre-trained language models (PLMs) have led to strong results in various domains including vision-language tasks~\cite{devlin2018bert,raffel2019exploring,brown2020language,radford2021learning}.
Such large PLMs can learn a new task with a few examples or generalize to a new task without fine-tuning on any training examples, i.e., few-shot and zero-shot learning~\cite{brown2020language,radford2021learning,tsimpoukelli2021multimodal}.
Few-shot learning overcomes the challenges of data-hungry supervised learning, where collecting human-labeled data is costly and slow. 
However, recent few-shot models such as GPT3~\cite{brown2020language}, Frozen~\cite{tsimpoukelli2021multimodal}, and PICa~\cite{yang2021empirical} are too large to deploy in small or moderate computing machines due to their gigantic model sizes

In this paper, we study low-resource learning of VL tasks with our proposed method, \method, a moderate-sized vision-language model, in which we fine-tune the model with no or a handful of training examples. 
For \method, we pre-train a sequence-to-sequence transformer model~\cite{cho2021unifying,raffel2019exploring} with prefix language modeling (PrefixLM) and masked language modeling (MaskedLM).
This setup is more practical in that training and inference can be run  economically using standard computing hardware
and it is expensive to obtain a large number of quality training examples in the real world.
In such a few-shot setting, task-specific prompts or task descriptions are important and have shown effectiveness in few-shot NLP tasks~\cite{gao2020making,radford2021learning,schick2020exploiting,schick2020s,brown2020language}.



To extend the success to VL tasks, we aim to answer the following questions for prompt-based low-resource VL learning.
Q1) How does prompt design affect zero/few-shot learning on new tasks? Q2) Does prompt design still matter given larger training? Q3) How do different pre-training objectives affect zero/few-shot learning?
To answer these questions, we explore various prompt formats including hand-crafted and noisy prompts on zero/few-shot VL learning datasets.
In addition, we study pre-training objectives on few-shot tasks inspired by \citet{raffel2019exploring}: prefix language modeling (PrefixLM) inspired by \citet{raffel2019exploring} and masked language modeling (MaskedLM).
To this end, we investigate the model's performance on few-shot VL tasks including visual question answering~\cite{goyal2017making,marino2019ok,hudson2019gqa}, captioning~\cite{agrawal2019nocaps,young2014image} (Fig.~\ref{fig:fewshot}), and miniImageNet~\cite{vinyals2016matching}.

In our empirical analysis, our \method\xspace with prompt-based learning outperforms Frozen \cite{tsimpoukelli2021multimodal} which is 31$\times$ larger than \method\xspace by 18.2\% point on zero-shot VQAv2 and achieves comparable results to a 246$\times$ larger model, PICa \cite{yang2021empirical}.
Furthermore, we observe that (1) prompts significantly affect zero-shot performance but marginally affect few-shot performance on new tasks (\S\ref{sec:zero} and \S\ref{sec:few}), (2) models with noisy prompts learn as quickly as hand-crafted prompts given larger training data (\S\ref{sec:exp:prompts}), and (3) MaskedLM helps few-shot VQA tasks while PrefixLM boosts captioning performance (\S\ref{sec:exp:obj}).



\section{Related Work}



\para{Vision-language few-shot learning.}
Recently, several few-shot learners on vision-language tasks were proposed including GPT~\cite{radford2019language,brown2020language},  Frozen~\cite{tsimpoukelli2021multimodal}, PICa~\cite{yang2021empirical}, and SimVLM~\cite{wang2021simvlm}.
Frozen~\cite{tsimpoukelli2021multimodal} is a large language model based on GPT-2~\cite{radford2019language}, and is transformed into a multimodal few-shot learner by extending the soft prompting to incorporate a set of images and text.
Their approach shows the few-shot capability on visual question answering and image classification tasks.
Similarly, PICa~\cite{yang2021empirical} uses GPT-3~\cite{brown2020language} to solve VQA tasks in a few-shot manner by providing a few in-context VQA examples. 
It converts images into textual descriptions so that GPT-3 can understand the images.
SimVLM~\cite{wang2021simvlm} is trained with prefix language modeling on weakly-supervised datasets.
It demonstrates its effectiveness on a zero-shot captioning task.
While these models achieve improvement on few-shot tasks, they are impractical to use in real-world applications due to their model sizes.

\para{Language model prompting.}
Providing prompts or task descriptions play an vital role in improving pre-trained language models in many tasks~\cite{gao2020making,radford2021learning,schick2020exploiting,schick2020s,brown2020language}.
Among them, GPT models~\cite{radford2019language,brown2020language} achieved great success in prompting or task demonstrations in NLP tasks.
In light of this direction, prompt-based approaches improve small pre-trained models in few-shot text classification tasks~\cite{gao2020making,schick2020exploiting,schick2020s}.
CLIP~\cite{radford2021learning} also explores prompt templates for image classification which affect zero-shot performance.
We follow these core ideas so we aim to improve zero-shot and few-shot performance using prompts in vision-language tasks.

\section{Analysis Setup}

\begin{figure}[tb!]
    \centering
    {\includegraphics[width=0.9\columnwidth]{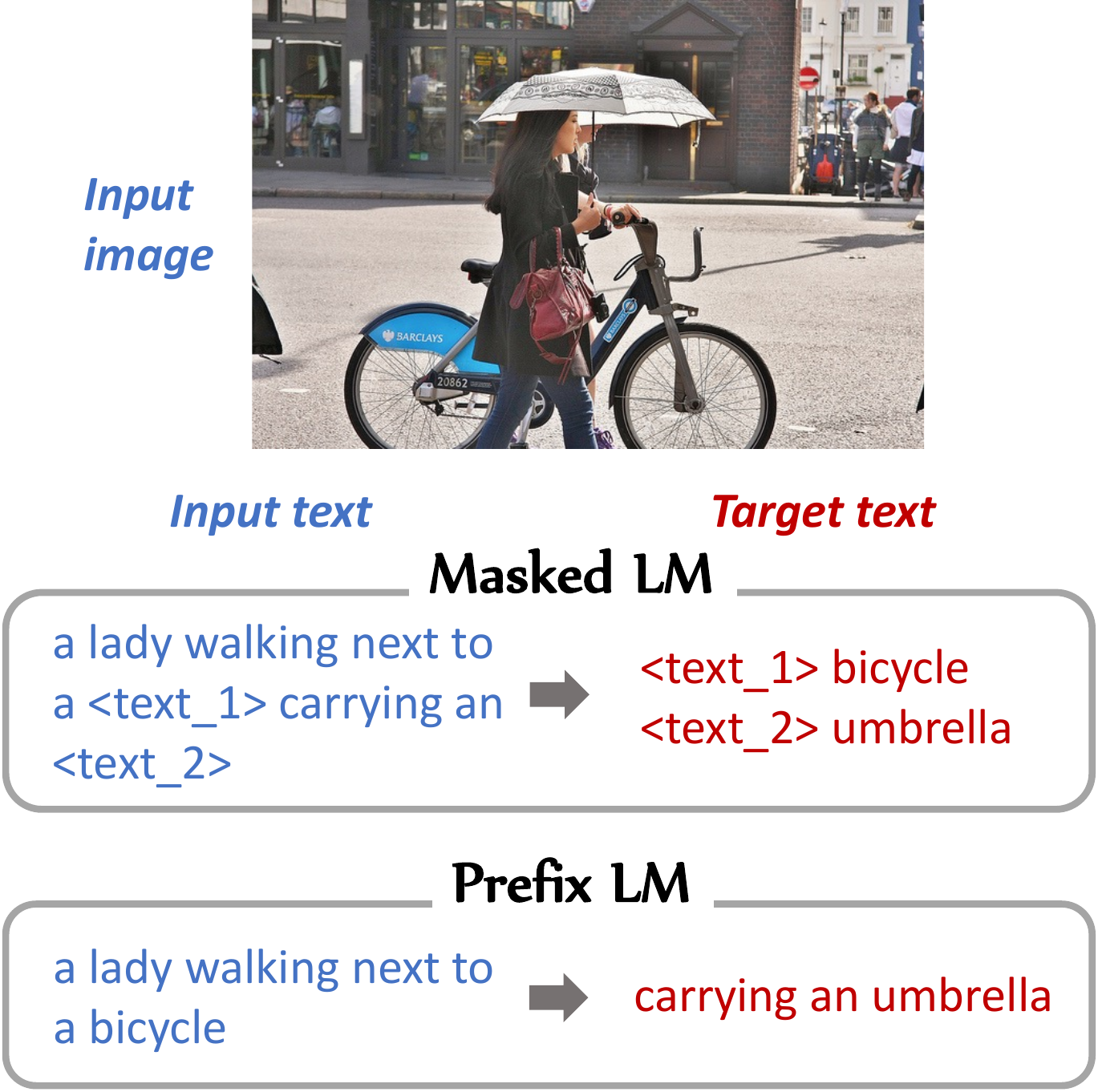}}
    \caption{\textbf{Pre-training objectives.}  We pre-train \method\xspace with masked language modeling (MaskedLM) and prefix language modeling (PrefixLM). 
    }
    \label{fig:obj}
\end{figure}

In this work, we study the zero-shot and few-shot performance of vision-language models $\mathcal{L}$.
We introduce our analysis setup: problem formulation, analysis questions, downstream tasks and datasets, evaluation metrics, and baselines. 



\subsection{Problem Formulation}
For zero-shot tasks, a pre-trained VL model $\mathcal{L}$ have no access to training set $\mathcal{D}_{train}$ and development set $\mathcal{D}_{dev}$, and directly makes inference on the test instances $\mathcal{D}_{test}$.
For few-shot tasks, we compose a dev set $\mathcal{D}_{dev}$ from training data and ensure that $|\mathcal{D}_{train}| =|\mathcal{D}_{dev}|$ following \citet{perez2021true,gao2020making} to tune the hyper-parameters and select the model.
We limit the sizes of training and development sets to meet the goal of learning from limited data.
The size of $\mathcal{D}_{train}$ and $\mathcal{D}_{dev}$ are small --- i.e., we set the size of both to 16 in our study.

\begin{table*}[tb!]
\centering
\caption{\textbf{Hand-crafted prompts.} We study hand-crafted prompts on zero-shot and few-shot tasks. \texttt{[Q]} and \texttt{[A]} refer to question text and answer text, respectively.  
\texttt{<text\_1>} is a sentinel token. We append image features to input text.
Target prompts are ``\texttt{[A]}'' and ``\texttt{<text\_1>} \texttt{[A]}'' in VQA. 
We use caption text as a target prompt in captioning. 
}
\resizebox{0.98\linewidth}{!}{
\begin{tabular}[t]{L{1.8cm}L{0.8cm}L{4.0cm}L{14.0cm}}
\toprule
    \textbf{Task} & \textbf{ID} &\textbf{Input prompt}  & \textbf{Example} \\
\midrule
\multirow{3}{*}{VQA}
                & P1 & \texttt{[Q]} \texttt{<text\_1>}   & \textbf{input:} What position is this man playing? \texttt{<text\_1>} \textbf{output:} \texttt{<text\_1>} pitcher   \\
\cmidrule(lr{1em}){2-4}
                & P2 & \textit{question:} \texttt{[Q]} \textit{answer:}   &  \textbf{input:} question: What position is this man playing? answer: \textbf{output:} \texttt{<text\_1>} pitcher    \\
\cmidrule(lr{1em}){2-4}
                & P3 & \textit{question:} \texttt{[Q]} \textit{answer:} \texttt{<text\_1>}    & \textbf{input:} question: What position is this man playing? answer: \texttt{<text\_1>} \textbf{output:} \texttt{<text\_1>} pitcher  \\
\midrule
\multirow{3}{*}{Captioning} 
    & Q1 & \textit{a picture of}  & \textbf{input:} a picture of \textbf{output:} a small black dog standing over a plate of food.   \\
\cmidrule(lr{1em}){2-4}
    & Q2 & \textit{a photo of}   &  \textbf{input:} a photo of \textbf{output:} a small black dog standing over a plate of food.  \\
\cmidrule(lr{1em}){2-4}    
    & Q3 & \textit{an image of} &  \textbf{input:} an image of \textbf{output:} a small black dog standing over a plate of food.  \\
\bottomrule
\end{tabular}}
\label{tab:prompts}
\end{table*}

\subsection{Analysis Questions}


We aim to answer the following questions in this study through experiments on multiple VL datasets.

\para{Q1) How does prompt design affect zero/few-shot learning on new tasks?}
Providing a pre-trained language model with task-specific prompts or significantly improves zero-shot and few-shot performance on NLP domains~\cite{gao2020making,schick2020exploiting,schick2020s,brown2020language}.
For this question, we test several ad-hoc prompts on vision-language tasks and analyze how large zero-shot and few-shot performance is affected by different prompts, hand-crafted and noisy prompts, in Sec.~\ref{sec:exp:prompts}.




\para{Q2) Does prompt design still matter given larger training data?}
As we will see in our experiments, prompts affect the zero/few-shot performance.
However, prompts may have different effects when models are given different sizes of training data.
To answer this question, we train models with different sizes of training data and various prompts, and compare the performance between different prompts.

\para{Q3) How do different pre-training objectives affect zero/few-shot performance?}
We study two different pre-training objectives on few-shot performance: prefix language modeling (PrefixLM) inspired by \citet{raffel2019exploring} and masked language modeling (MaskedLM).
In this setup, we pre-train our model with different objectives and test the model on zero-shot and few-shot tasks in Sec.~\ref{sec:exp:obj}.

\subsection{Downstream Tasks and Datasets}
In this work, we mainly focus on three tasks: visual question answering, captioning, and categorical learning.
The visual question answering task requires models to answer a question to a given context image.
We convert the visual question answering task into a generation task so that the model can generate answers in the zero-shot setting.
The captioning task requires a model to generate descriptions for a given context image.
The categorical learning requires a model to choose the correct category or class.
We evaluate our model in an open-ended fashion to quantify fast learning of categories, in which it must generate correct labels unlike other classification methods.

We include VQAv2~\cite{goyal2017making}, OK-VQA~\cite{marino2019ok}, and GQA~\cite{hudson2019gqa} for visual question answering tasks, and NoCaps~\cite{agrawal2019nocaps}, and Flickr30k~\cite{young2014image} for image captioning.\footnote{We include COCO captioning results on Sec.~\ref{app:coco} of Appendix.}
We use Karpathy split~\cite{karpathy2015deep} for Flickr30k, which re-splits train and val images into 29,000 / 1,014 / 1,000 for train / validation / test.
For categorical learning, we include miniImageNet~\cite{vinyals2016matching}, a meta learning dataset.
Following \cite{tsimpoukelli2021multimodal}, we use only meta test data to evaluate \method\xspace in a few-shot manner and test on 5-way $k$-shot setup, where 5 classes and $k$ examples \textit{per class} are given.\footnote{For VQA and captioning, we include $k$ samples in total, not per class.}

\subsection{Evaluation Metrics}
To evaluate few-shot performance, we randomly sample 5 different training and dev splits and measure average performance on the 5 splits. 
We fine-tune the vision-language models with 200 epochs for the few-shot setup and choose the best checkpoint on the dev set.
For NoCaps task, it does not have training data. Thus we use the training data from COCO captioning in the experiments following~\citet{wang2021simvlm}. 
We evaluate on the VQAv2 validation set, GQA test-dev, OK-VQA test set, test set of Karpathy split for Flickr30k captioning, and NoCaps validation set.
We adopt accuracy for VQA datasets and miniImageNet, and CIDEr~\cite{vedantam2015cider} and SPICE~\cite{anderson2016spice} as evaluation metrics for captioning.

\subsection{Baselines}
We evaluate strong zero/few-shot vision-language learners for comparison: Frozen~\cite{tsimpoukelli2021multimodal}, PICa~\cite{yang2021empirical} for VQA datasets and SimVLM~\cite{wang2021simvlm} for captioning datasets.
We include Unified VLP~\cite{zhou2020unified} for few-shot VQAv2 and Flickr30k.
Also, we compare them with fully fine-tuned models $\mathcal{L}_{full}$ as upper bounds of few-shot models for each task; these models are fine-tuned on the entire datasets while few-shot models can access a small amount of data.
For fully fine-tuned models $\mathcal{L}_{full}$, we borrow numbers from Uniter$_{large}$~\cite{chen2019uniter} for VQAv2, Oscar~\cite{li2020oscar} for GQA, SimVLM~\cite{wang2021simvlm} and VinVL~\cite{zhang2021vinvl} for NoCaps CIDER and SPICE respectively, and Unified VLP~\cite{zhou2020unified} for Flickr30k captioning. 
We include VL-T5$_\text{no-vqa}$ as a baseline which is pre-trained without visual question answering datasets~\cite{cho2021unifying}.
For miniImageNet, we include Frozen and AFHN~\cite{li2020adversarial}. Frozen is designed for few-shot learning while AFHN is for meta learning, which is smaller and faster.

\section{Method}
%


Before diving into the analysis, we introduce our model, \method, to do zero/few-shot learning on VL tasks and answer the analysis questions we raised.
We introduce \method\xspace architecture and pre-training objectives.



\subsection{Encoder-decoder Vision-language Model}
We adopt an encoder-decoder architecture~\cite{cho2021unifying,vaswani2017attention}, to encode visual and text inputs and generate target text.
We represent an input image with 36 object regions from a Faster R-CNN~\cite{ren2015faster} trained on Visual Genome~\cite{krishna2017visual}.
The sets of region representations are fed into the encoder by appending them to the text \citet{cho2021unifying}.
We train the model parameters $\theta$ by minimizing the negative log-likelihood of target text $y$ tokens given input text $x$ and image $v$:
\begin{equation}
\label{loss}
L_\theta = -\sum_{i=1}^{|y|} \log P_\theta (y_i | y_{<i}, x, v).
\end{equation}
The model is not task-specific, so it is a  good option for zero/few-shot settings.


\subsection{Pre-training Objectives}
We pre-train the models with both prefix language modeling (PrefixLM) and masked language modeling (MaskedLM).
Fig.~\ref{fig:obj} illustrates the PrefixLM and MaskedLM.

\para{Prefix language modeling.}
We include prefix language modeling (PrefixLM) following \citet{raffel2019exploring}.
Given an image and a span of text, this objective randomly splits the text into two separate components; the former component with the given image is used as inputs to the encoder and the latter component is used as target text to be generated by the decoder.

\para{Masked language modeling.}
We follow \citet{cho2021unifying} to do masked language modeling.
This objective is to replace random spans with numbered sentinel tokens, e.g., \texttt{<text\_1>}, and then the masked text is fed into the encoder.
Then the decoder generates the masked spans as target text.
We randomly mask 15\% of input text tokens and replace them with sentinel tokens.

\para{Pre-training data.}
To pre-train \method, we collect image-caption data from MS COCO~\cite{lin2014microsoft,chen2015microsoft} and Visual Genome (VG)~\cite{krishna2017visual}.
The pre-training datasets contains 9.18M image-text pairs and 180K distinct images.

\begin{table*}[tb!]
\parbox{.48\linewidth}{
\centering
\caption{\textbf{Zero-shot VQA results.} We test models without any training examples. VL-T5$_\text{no-vqa}$ is pre-trained without VQA datasets. Compared to larger models, Frozen and PICa-Full, our models outperform them or show the comparable results. 
}
\resizebox{\linewidth}{!}{
\begin{tabular}[t]{L{3cm}|C{1.5cm}|C{1.5cm}C{1.5cm}C{1.5cm}}
\toprule
    \textbf{Model} &\textbf{Model size}   &\textbf{VQAv2}   &\textbf{OK-VQA}    &\textbf{GQA}     \\
\midrule
Unified VLP & 122M &0.0 &-   &- \\
VL-T5$_\text{no-vqa}$ &224M &13.5   &5.8   &6.3   \\
Frozen &7B &29.5   &5.9  & -  \\
PICa &175B &-   &\textbf{17.5}   &-    \\
\midrule
\method$_{base}$  &224M    &43.4   &11.6   &27.0     \\
\method$_{large}$  &740M   &\textbf{47.7}   &\textit{16.5}   &\textbf{29.3}    \\
\bottomrule
\end{tabular}}

\label{tab:zeroshotvqa}
}
\hfill
\parbox{.48\linewidth}{
\centering
\caption{\textbf{Few-shot VQA results.} We report average performance over 5 different splits. The size of training and validation sets are 16 for our \method\xspace and VL-T5$_\text{no-vqa}$, and Frozen and PICa use 4 and 16 in-context training examples, respectively. For the fair comparison to Frozen, we include \method$_{base}^*$ with 4 training and validation examples. 
}
\resizebox{\linewidth}{!}{
\begin{tabular}[t]{L{3cm}|C{1.5cm}|C{1.5cm}C{1.5cm}C{1.5cm}}
\toprule
    \textbf{Model} &\textbf{Model size}   &\textbf{VQAv2}   &\textbf{OK-VQA}    &\textbf{GQA}     \\

\midrule
Unified VLP & 122M & 24.3   &-   & -\\
VL-T5$_\text{no-vqa}$ &224M &31.8   &12.7   &19.6   \\
Frozen   &7B &38.2   &12.6  & -  \\
PICa  &175B &\textbf{54.3}   &\textbf{43.3}   &-    \\
\midrule
\method$_{base}^*$  &224M   &45.1   &14.5   &26.9   \\
\method$_{base}$  &224M   &48.2   &15.0   &32.2   \\
\method$_{large}$  &740M   &\textit{51.1}  &\textit{23.1}   &\textbf{35.7}      \\
\midrule
Fine-tuned $\mathcal{L}_{full}$  &-   & 72.6    & -  &61.5     \\
\bottomrule
\end{tabular}}
\label{tab:fewshotvqa}
}
\end{table*}

\begin{table*}[tb!]
\parbox{.48\linewidth}{
\centering
\caption{\textbf{Zero-shot captioning results.}  We use the CIDEr and SPICE metrics for evaluation. 
}
\resizebox{\linewidth}{!}{
\begin{tabular}[t]{L{2.7cm}|C{1.5cm}|C{1.2cm}C{1.2cm}C{1.2cm}C{1.2cm}}
\toprule
\multicolumn{1}{c|}{\multirow{2}{*}{\hspace{-1.3cm}\textbf{Model}}} & \multicolumn{1}{c|}{\multirow{2}{*}{\textbf{Model size}}} &  \multicolumn{2}{c}{\textbf{NoCaps}} & \multicolumn{2}{c}{\textbf{Flickr30k}}   
\\ \cmidrule(lr){3-4} \cmidrule(lr){5-6} 
&   &\textbf{CIDEr} &\textbf{SPICE}   &\textbf{CIDEr}    &\textbf{SPICE}    \\ 
\midrule
Unified VLP & 122M  & -  &-   &24.9   &7.2  \\
VL-T5$_\text{no-vqa}$ &224M    &4.4 &5.3   &2.6    &2.0 \\
SimVLM$_{huge}$  &-     &\textbf{101.4} &-  &-   &-  \\
\midrule
\method$_{base}$  &224M     &42.2   &8.5  &31.0    &10.0   \\
\method$_{large}$  &740M      &\textit{47.7}    &\textbf{9.1} &\textbf{36.5}    &\textbf{10.7}   \\
\bottomrule
\end{tabular}}
\label{tab:zeroshotcaption}
}
\hfill
\parbox{.48\linewidth}{
\centering
\caption{\textbf{Few-shot captioning results.} We report average performance over 5 different splits. We use the CIDEr and SPICE metrics for evaluation. 
}
\resizebox{\linewidth}{!}{
\begin{tabular}[t]{L{2.7cm}|C{1.5cm}|C{1.2cm}C{1.2cm}C{1.2cm}C{1.2cm}}
\toprule
\multicolumn{1}{c|}{\multirow{2}{*}{\hspace{-1.3cm}\textbf{Model}}} & \multicolumn{1}{c|}{\multirow{2}{*}{\textbf{Model size}}} &  \multicolumn{2}{c}{\textbf{NoCaps}} & \multicolumn{2}{c}{\textbf{Flickr30k}}   
\\ \cmidrule(lr){3-4} \cmidrule(lr){5-6} 
&   &\textbf{CIDEr} &\textbf{SPICE}   &\textbf{CIDEr}    &\textbf{SPICE}    \\ 

\midrule
Unified VLP & 122M  &-   &-   &28.8   &9.4  \\
VL-T5$_\text{no-vqa}$ &224M    &22.0    &6.8  &12.8    &8.3 \\
\midrule
\method$_{base}$  &224M   &48.6 &10.0    &32.6  &12.8   \\
\method$_{large}$  &740M      &\textbf{53.1}    &\textbf{10.4}  &\textbf{37.0}&\textbf{13.5}   \\
\midrule
Fine-tuned $\mathcal{L}_{full}$  &-  &112.2    &13.1 &67.4&17.0   \\
\bottomrule
\end{tabular}}
\label{tab:fewshotcaption}
}
\end{table*}

\section{Low-resource Adaptation}


In downstream tasks, we train our model with few-shot examples.
Fig.~\ref{fig:illust} shows an illustration of \method\xspace in inference time.
Given a prompt template $\mathcal{P}$, we first get input text and target text using the template $x,y = \mathcal{P}(\text{input},\text{label})$.
Then we train model parameters by minimizing the negative log-likelihood in Eq.~\eqref{loss}.
In inference, we use the same prompt and the model generates the label text. 
Here we obtain the final label by removing the target prompt template.


\subsection{Prompt Design}


Prompts affect the performance of the vision-language model~\cite{cho2021unifying}; we study the effect of different prompts on the zero-shot and few-shot performance on downstream tasks.
Tables~\ref{tab:prompts} and \ref{tab:prompts2} show prompts we used in our experiments.

\subsubsection{Visual Question Answering.}
The visual question answering tasks (VQA, OK-VQA, and GQA) require models to answer a question to a given context image.
Recent approaches~\cite{chen2019uniter,tan2019lxmert,su2019vl,li2019visualbert,li2020oscar} tackle visual question answering tasks as multi-label classification over a predefined set of answer candidates. 
Instead, we approach the visual question answering tasks as a generation task so that the model can produce the answers without introducing any task-specific heads.
In this setup, prompts act as constraints to guide the models to generate proper formats of answers;  models might generate a sentence for VQA, which is not the correct format, without prompts.  

Therefore, we study several prompts for input and output as shown in Tables~\ref{tab:prompts} and \ref{tab:prompts2};
we explore hand-crafted prompts (Table~\ref{tab:prompts}) and noisy prompts for ablation study (Table~\ref{tab:prompts2}).

\para{Hand-crafted prompts.}
For input prompts, we explore three different templates: ``\textit{question:} \texttt{[Q]} \textit{answer:}'' and with the \texttt{<text\_1>} sentinel token at the end.
Similarly to masked language modeling, we expect models to generate words thanks to the sentinel token.
For target prompts, we explore two different templates: ``\texttt{[A]}'' (an answer) and ``\texttt{<text\_1> [A]}'' (an answer with a sentinel token). 
Here, we aim to mimic MaskedLM's target text format, so the similar format helps the model quickly adapt to the new task.
We call each prompt ID as in Table~\ref{tab:prompts}.


\para{Noisy prompts.}
To understand the effect of noisy prompts in zero/few-shot learning, we include irrelevant prompts, noisy tokens, and random sentences as in Table~\ref{tab:prompts2}.
Irrelevant prompts are random questions or instructions that mislead models to answer wrong questions or follow irrelevant instructions. 
Noisy tokens are randomly selected from T5's vocabulary, so we test how robust our model is to random tokens.
Finally, random sentences are captions from MS COCO and this gives false information to models.

\subsubsection{Captioning.}
In NoCaps and Flickr30k, we explore three hand-crafted input prompts: ``\textit{a picture of}'', ``\textit{a photo of}'', and ``\textit{an image of}''. 
We study the effect of different word choices in this captioning task. While the three different words have similar meanings, they show different performance in zero-shot and few-shot tasks as we will see in our experiments..
For target prompts, we just train the model with the original caption without any additional prompts. 


\subsubsection{MiniImageNet}
In miniImageNet, we train our model with a hand-crafted input prompt, ``\textit{This is} \texttt{<text\_1>},'' and target prompt, ``\texttt{<text\_1>} \texttt{[A]}.''
We compare our model with and without prompts in this dataset to study whether prompts are helpful in categorical learning.





\section{Results and Discussion}
In this section, we first discuss our main results on zero-shot and few-shot tasks and then answer the questions we raised: does prompt design matter in zero/few-shot learning?

\subsection{Experiment Details} 
For pre-training, we set batch size 1,280 and 800 for \method$_{base}$ and \method$_{large}$, respectively and pre-train them with 30 epochs. 
We use learning rate 1e-4 with 5\% linear warmup.
For few-shot learning, we train models with 200 epochs, learning rate 5e-5 and 5\% linear warmup and choose the best checkpoint on the dev set.
For \method, we use ``question: \texttt{[Q]} answer \texttt{<text\_1>}'' (P3) as an input prompt and ``\texttt{<text\_1>} \texttt{[A]}'' as a target prompt for visual question answering, and ``an image of'' (Q3) as an input prompt for captioning, which show the best performance. 
We will study the effect of different prompts in Sec.~\ref{sec:exp:prompts}.
The sizes of of $\mathcal{D}_{train}$ and $\mathcal{D}_{dev}$ are 16 on VQA and captioning tasks.
For miniImageNet, we use `This is \texttt{<text\_1>},'' and ``\texttt{<text\_1>} \texttt{[A]}'' as input and target prompts.
In this data, we test with \{1, 3, 5\}-shots per class.

\subsection{Performance on Zero-shot Learning}
\label{sec:zero}
We evaluate the existing models in a zero-shot manner, in which models do not have access to any training data.
Tables~\ref{tab:zeroshotvqa} and \ref{tab:zeroshotcaption} show the results on VQA and captioning datasets, respectively.
First, \method\xspace with the hand-crafted prompt (P3) achieves better performance than other baselines on VQA datasets.
In particular, our \method$_{base}$ significantly outperforms Frozen which is about 31$\times$ larger than ours.
Also, PICa based on GPT3~\cite{brown2020language} shows the best performance on OK-VQA. 
It is noticeable that our \method$_{large}$, the 246$\times$ smaller model, achieves the comparable result to PICa.
Compared to VL-T5$_\text{no-vqa}$ which is the same architecture as ours, \method$_{base}$ improves VQAv2 performance by about 30\% point. 
As we will see in the later section, our pre-training objectives and the prompts boost the VQA performance. 
On NoCaps, SimVLM$_{huge}$ shows the best performance.
Our \method$_{base}$ significantly improves the performance compared to VL-T5$_\text{no-vqa}$.
As we will see in the later section, our pre-training objectives and the prompts boost the VQA and captioning performance. 


\begin{table}[tb!]
\centering
\caption{\textbf{5-way miniImageNet results.} We evaluate \method\xspace in a generative manner. The shot represents the number of training examples per class.}
\resizebox{0.98\linewidth}{!}{
\begin{tabular}[t]{L{3cm}|C{1.5cm}|C{1.5cm}C{1.5cm}C{1.5cm}}
\toprule
    \textbf{Model} &\textbf{Model size}   &\textbf{1 shot }   &\textbf{3 shots}    &\textbf{5 shots}     \\
\midrule
Frozen   &7B & 14.5   & 34.7   & 33.8   \\
\midrule
\method$_{base}$ (no prompt)  &224M   &48.0  &75.0    & 82.6   \\
\method$_{base}$  &224M   & 57.0   &78.0    & 84.2   \\
\method$_{large}$  &740M   & 57.1   & 78.3   & 84.4      \\
\midrule
AFHN  & - & 62.3   & -   & 78.1      \\
\bottomrule
\end{tabular}}
\label{tab:fewshotimage}
\end{table}

\subsection{Performance on Few-shot Learning}
\label{sec:few}


Tables~\ref{tab:fewshotvqa} and  \ref{tab:fewshotcaption} show the few-shot performance on VQA and captioning datasets.
Sizes of training and validation sets are 16 for \method, VL-T5$_\text{no-vqa}$, and Unified VLP; and Frozen and PICa use 4 and 16 in-context demonstration examples, respectively.
 
On VQAv2 and OK-VQA, PICa shows the best performance while our \method$_{large}$ achieves the comparable result on VQAv2.
OK-VQA requires external knowledge to answer unlike other VQA datasets, so larger models and large pre-training data (prior knowledge) are necessary to improve.
Interestingly, \method$_{base}^*$, which is trained with 4 training examples, outperforms Frozen.
On captioning data, \method$_{base}$ notably outperforms VL-T5$_\text{no-vqa}$ by 31.1\% point on NoCaps CIDEr.

Unified VLP slightly underperforms \method\xspace on Flickr30k captioning task.
We conjecture that their architecture is based on a encoder-decoder transfomer and it is pre-trained with a captioning task~\cite{zhou2020unified}.


\subsection{MiniImageNet}
Table~\ref{tab:fewshotimage} shows results on miniImageNet, where models must choose the correct class for each image.
We train and evaluate \method\xspace in an generative manner; the model must generate correct label text to get the credit.
\method\xspace significantly outperforms Frozen in all shots. Note that we train \method\xspace with a few training samples while Frozen uses them as in-context demonstration.
Interestingly, \method\xspace with a hand-crafted prompt improves performance a lot on the 1-shot case, while it marginally improves on the 5-shot case.

\begin{table}[tb!]
\centering
\caption{\textbf{Zero-shot results of hand-crafted prompts.} We test different input prompts in zero-shot predictions.
We use a CIDEr metric for Flickr30k.
Note that zero-shot setting does not require target prompts.
}
\resizebox{0.98\linewidth}{!}{
\begin{tabular}[t]{L{1.7cm}|C{1.8cm}C{1.7cm}C{1.7cm}C{1.7cm}}
\toprule
    &  \textbf{no prompt}  &  \textbf{P1}  &  \textbf{P2}  &  \textbf{P3}   \\
\midrule
\textbf{VQAv2} &3.7   &9.9   &19.0    &43.4    \\
\midrule
  &  \textbf{no prompt}  & \textbf{Q1}   & \textbf{Q2}  & \textbf{Q3}   \\
 \midrule
\textbf{Flickr30k} &9.6   &15.2   &25.6    &31.0 \\
\bottomrule
\end{tabular}}
\label{tab:promptvqa}
\end{table}


\subsection{Study of Prompt Design}
\label{sec:exp:prompts}

\begin{figure}[tb!]
    \centering
    \includegraphics[width=0.98\columnwidth]{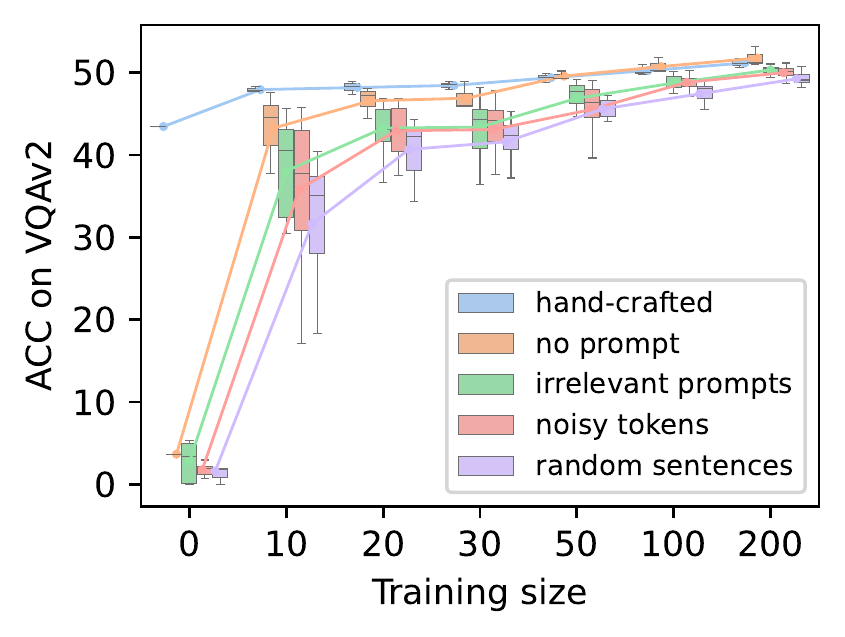}
    \caption{\textbf{VQAv2 results on noisy prompts.} We investigate different prompts on various training sizes. \method\xspace is trained with our best hand-crafted prompt (P3), irrelevant prompts, noisy tokens and random sentences. We list the prompt templates in Table~\ref{tab:prompts2} of appendix. We use ``\texttt{<text\_1>} \texttt{[A]}'' as our target prompt.}
    \label{fig:prompt}
\end{figure}
\begin{figure}[tb!]
    \centering
    \includegraphics[width=0.98\columnwidth]{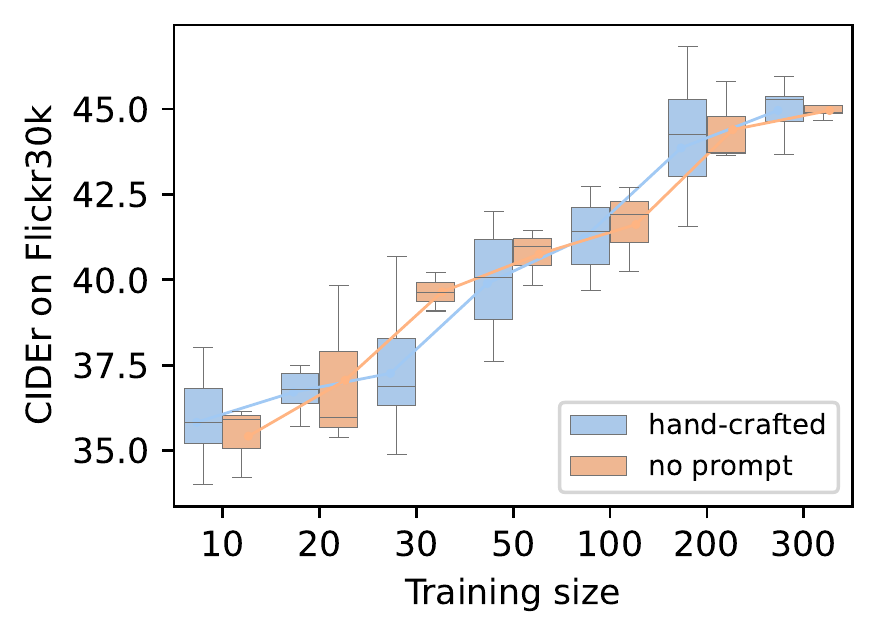}
    \caption{\textbf{Flickr30k results on hand-crafted prompts.}   We investigate different hand-crafted prompts (Q1, Q2, and Q3) on various training sizes. 
    }
    \label{fig:handpromptflickr}
\end{figure}
\begin{figure}[tb!]
    \centering
    \includegraphics[width=0.98\columnwidth]{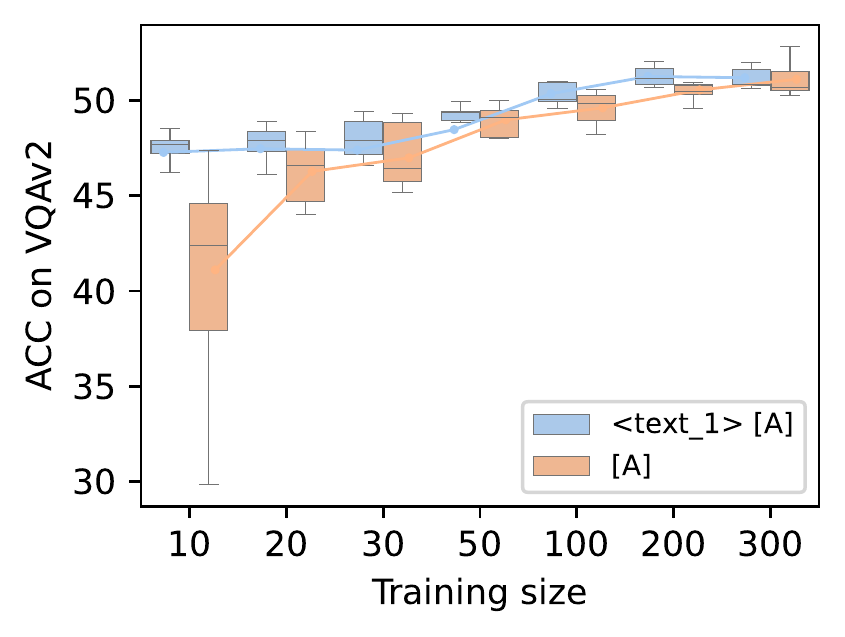}
    \caption{\textbf{VQAv2 results on different target prompts.}  We investigate different target prompts with hand-crafted input prompts on various training sizes. 
    }
    \label{fig:handpromptvqa}
\end{figure}

Here we examine the effect of different prompts on \method$_{base}$ in Table~\ref{tab:promptvqa} and Figs.~\ref{fig:handpromptvqa}, \ref{fig:handpromptflickr}, and \ref{fig:prompt}.
We test the model on VQAv2 and Flickr30k datasets.

\subsubsection{Zero-shot Predictions}
Table~\ref{tab:promptvqa} shows the zero-shot performance on VQAv2 and Flickr30k.
We observe that zero-shot results are remarkably affected by input prompts on both datasets.
For input prompts, \texttt{<text\_1>} in P1 and P3 helps the zero-shot predictions significantly compared to ``no prompt'' and P2.
We conjecture that \texttt{<text\_1>} guides the model to predict masked spans similarly to MaskedLM, so it improves the performance.

On Flickr30k, we examine different word choices of prompts: ``a picture of'' (Q1), ``a photo of'' (Q2), and ``an image of'' (Q3).
For instance, using ``an image of'' outperforms using no prompt by 21.4 point.
It is noticeable that different word choices significantly affect the zero-shot results.

\subsubsection{Few-shot Predictions}
We study various input prompts including irrelevant prompts, noisy tokens, and random sentences on VQAv2 (Fig.~\ref{fig:prompt}).
First, noisy prompts and no prompt achieve near 0 accuracy on the zero-shot setting.
In few-shot predictions, \method\xspace with noisy prompts learns as quickly as hand-crafted prompts given larger data.
For example, our model with noisy prompts achieves comparable results to the best hand-crafted prompt.
Among all different types of noisy prompts, random sentences deteriorate performance the most.
This is because the random sentences come from captions in MS COCO, so the model might choose the answer from wrong captions not from images.
Interestingly, no prompt outperforms the other noisy prompts and even shows similar to or better than the hand-crafted prompt with larger training data.
We also observe a similar phenomenon on Flickr30k; no prompt performs similar to hand-crafted prompts in Fig.~\ref{fig:handpromptflickr}.

In addition, we explore two different target prompts, ``\texttt{<text\_1} \texttt{[A]}'' and ``\texttt{[A]}.''
We try to mimic the MaskedLM's target text format, so we add ``\texttt{<text\_1}'' to target prompt on VQA.
This might help the model's fast adaptation to a new task since they share the same target prompt.
In Fig.~\ref{fig:handpromptvqa}, we notice an interesting phenomenon; the target prompt ``\texttt{[A]}'' shows a larger variance than the other suggesting that introducing ``\texttt{<text\_1}'' helps the model quickly adapt to a new task.
However, both prompts show similar results given larger training data, e.g., 300.

\begin{table}[tb!]
\centering
\caption{\textbf{Results on different pre-training objectives.} We test our pre-training objectives to investigate how it affects zero-shot and few-shot performance. We train \method$_{base}$ with 16 training and validation examples.
}
\resizebox{0.98\linewidth}{!}{
\begin{tabular}[t]{L{4cm}|C{1.5cm}C{1.5cm}C{1.5cm}C{1.5cm}}
\toprule
    \textbf{Objective} &\textbf{VQAv2}     &\textbf{GQA}   &\textbf{Flickr30k CIDEr}\\
\midrule
\textbf{Zero-shot}  &   &   &   \\
MaskedLM   &42.4   &25.1   &4.6   \\
PrefixLM   &11.9   &6.7   &26.8  \\
MaskedLM + PrefixLM    &\textbf{43.4}   &\textbf{27.0}   &\textbf{31.0}   \\
\midrule
\textbf{Few-shot}  &   &   &   \\
MaskedLM   &46.0   &31.4   &18.5   \\
PrefixLM   &40.8   &27.6   &31.8  \\
MaskedLM + PrefixLM    &\textbf{48.2}   &\textbf{32.2}   &\textbf{32.6}   \\
\bottomrule
\end{tabular}}
\label{tab:obj}
\end{table}


\subsection{Pre-training Objectives}
\label{sec:exp:obj}


We investigate how pre-training objectives affect different tasks. 
We pre-train \method\xspace with different pre-training objectives: masked language modeling (MaskedLM) and prefix language modeling (PrefixLM).

In Table~\ref{tab:obj}, we observe that MaskedLM helps VQA tasks while PrefixLM helps captioning tasks in zero-shot and few-shot settings. 
We conjecture that MaskedLM is to predict spans, which is analogous to predict correct answers to questions, and PrefixLM is to generate the rest of the given prefix, which is similar to captioning tasks.
In other words, if the pre-training task is similar to the downstream tasks, then it will help performance further.
When pre-training with both objectives, they create a synergetic effect and thus improve cross-task generalization.

\section{Conclusion}
In this work, we present \method, a few-shot prompt-based learner on vision-language tasks.
On diverse datasets, \method\xspace outperforms baselines and shows comparable results to PICa which is 246$\times$ larger than ours.
We observe that prompts are vital in zero-shot and few-shot tasks and each pre-training objective helps different few-shot tasks. 
Also, we find out that models with larger training data are not significantly affected by noisy prompts.
Future work includes exploring automatic prompt generation and diverse formats of few-shot tasks such as multiple-choice VQA.
Finding optimal prompts require exhaustive engineering to achieve the best performance and leads to impressive results. 
We leave the exploration of these directions to future investigations.


\bibliography{bibtex}
\bibliographystyle{acl_natbib}


\clearpage
\appendix


\begin{table}[tb!]
\centering
\caption{\textbf{Model architectures.} }
\resizebox{0.98\linewidth}{!}{
\begin{tabular}[t]{L{3cm}|C{2.5cm}C{2.5cm}}
\toprule
    \textbf{Hyperparameter}  &\textbf{\method$_{base}$}     &\textbf{\method$_{large}$} \\
\midrule
\# Layers &12+12   &24+24          \\
Hidden dimension &768   &1,024    \\
FF hidden size &3,072   &4,096    \\
\# Attention head  &12   &16   \\
Attention head size &64  &64    \\
\bottomrule
\end{tabular}}
\label{tab:modelpara}
\end{table}

\section{Model Architectures}
Table~\ref{tab:modelpara} shows model parameters in our model, \method. \method$_{base}$ and  \method$_{large}$ is based on VL-T5~\cite{cho2021unifying} and T5~\cite{raffel2019exploring}, respectively.

\begin{table}[]
    \centering
    \caption{\textbf{COCO captioning results.}  We use the CIDEr and SPICE metrics for evaluation. 
    }
    \resizebox{\linewidth}{!}{
    \begin{tabular}[t]{L{2.7cm}|C{1.5cm}|C{1.2cm}C{1.2cm}C{1.2cm}C{1.2cm}}
    \toprule
    \multicolumn{1}{c|}{\multirow{2}{*}{\hspace{-1.3cm}\textbf{Model}}} & \multicolumn{1}{c|}{\multirow{2}{*}{\textbf{Model size}}} &  \multicolumn{2}{c}{\textbf{Zero-shot}} & \multicolumn{2}{c}{\textbf{Few-shot}}   
    \\ \cmidrule(lr){3-4} \cmidrule(lr){5-6} 
    &   &\textbf{CIDEr} &\textbf{SPICE}   &\textbf{CIDEr}    &\textbf{SPICE}    \\ 
    \midrule
    VL-T5$_\text{no-vqa}$ &224M    &4.9 &2.0   &43.0    &10.8 \\
    SimVLM$_{huge}$  &-     &102.3 &22.1  &-   &-  \\
    \midrule
    \method$_{base}$  &224M     &84.5   &16.2     & 98.7  &18.9   \\
    \method$_{large}$  &740M      &92.1    &17.3 &100.4    &19.1   \\
    \midrule
    Unified VLP (fully supervised)  &122M   &-    &- &117.7 &21.3   \\
    \bottomrule
    \end{tabular}}
    \label{tab:cococap}
\end{table}

\section{COCO Captioning}
\label{app:coco}
We evaluate our model with COCO captioning data. We use Karpathy split~\cite{karpathy2015deep} for MS COCO captioning, which re-splits train and val images into 113,287 / 5000 / 5000 for train / validation / test.
Table~\ref{tab:cococap} shows the results on COCO.

\begin{table*}[tb!]
\centering
\caption{\textbf{Prompt templates.} We test different input prompts on VQAv2. \texttt{[Q]} refers to input question text.  We use \texttt{<text\_1> [A]} as target text. We append image features to input text. 
}
\resizebox{0.9\linewidth}{!}{
\begin{tabular}[t]{L{14cm}L{3cm}}
\toprule
    \textbf{Input prompt template} &  \textbf{Category} \\
\midrule
    Fill in the blank in the below sentence: \texttt{[Q]} & irrelevant prompts  \\
    Question: \texttt{[Q]} True or False?   & irrelevant prompts   \\
    \texttt{[Q]}  What color is the floor?  & irrelevant prompts   \\
    Paraphrase this into a different question? \texttt{[Q]}    & irrelevant prompts   \\
    \texttt{[Q]} How many are they?    & irrelevant prompts   \\
\midrule
    nezg publice passed Dream \texttt{[Q]}    & noisy tokens \\
    benefic video starting garbagetap Talent summary \texttt{[Q]}    & noisy tokens \\
    gestion Bun dates youngest batteriesfeder organisationoyez \texttt{[Q]}    & noisy tokens \\
    \texttt{[Q]} chefernției geekutilisées plantingasta Pest principiiMF saddle véritable   & noisy tokens \\
    \texttt{[Q]} composant emergency laissé Klägereiniger swipe concentrateOSS/18 rewardprepaid   & noisy tokens \\
\midrule
    \texttt{[Q]} A black dog is sitting on a couch. & random sentences \\
    \texttt{[Q]} A man working at a kitchen counter in a room illuminated by sunlight.   & random sentences \\
    A brown purse is sitting on a green bench. \texttt{[Q]}    & random sentences \\
    A television that is sitting next to signs. \texttt{[Q]}    & random sentences \\
    \texttt{[Q]} A woman is wearing white pants.    & random sentences \\
\bottomrule
\end{tabular}}
\label{tab:prompts2}
\end{table*}

\begin{figure}[tb!]
    \centering
    \includegraphics[width=0.98\columnwidth]{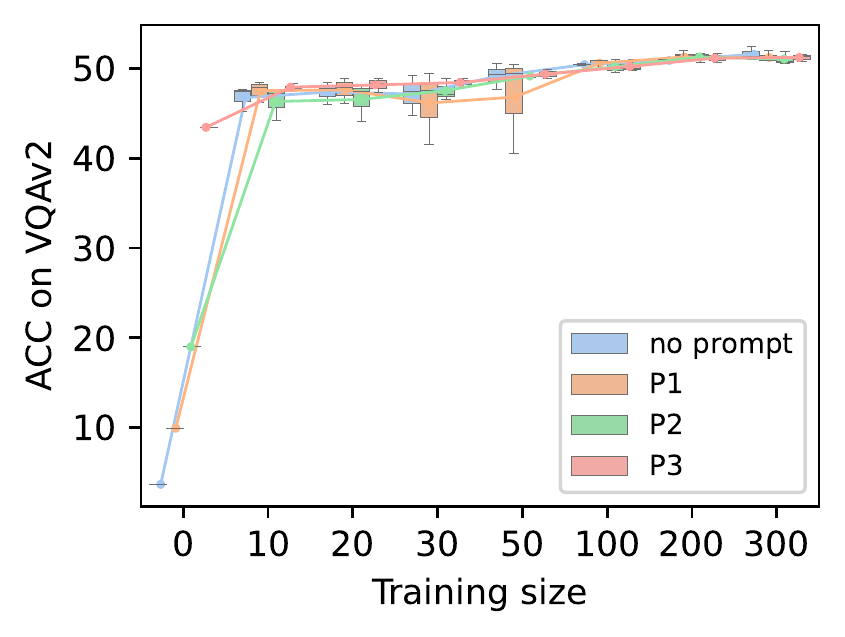}
    \caption{\textbf{VQAv2 results on hand-crafted prompts and the target prompt ``\texttt{<text\_1>} \texttt{[A]}''.}  }
    \label{fig:handpromptvqa_detail}
\end{figure}

\begin{figure}[tb!]
    \centering
    \includegraphics[width=0.98\columnwidth]{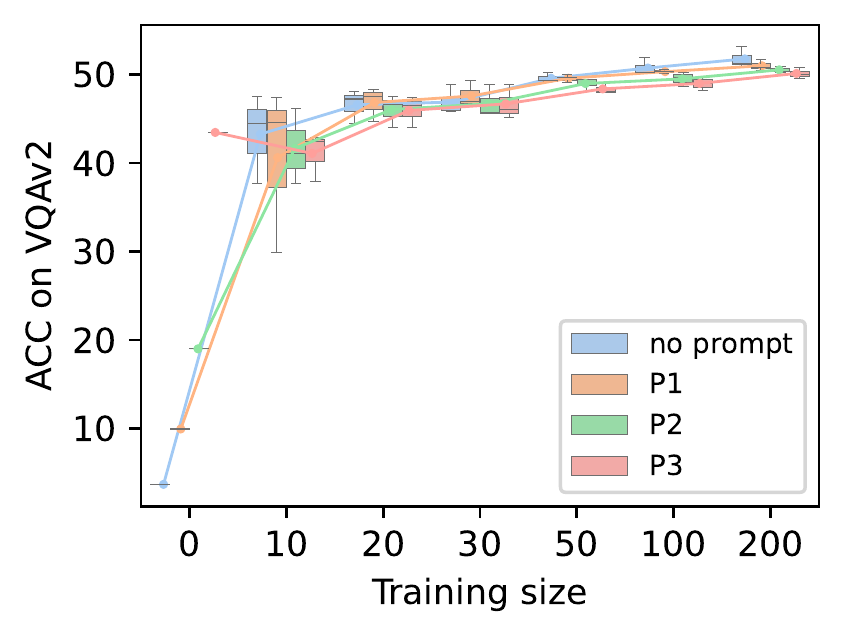}
    \caption{\textbf{VQAv2 results on hand-crafted prompts and the target prompt ``\texttt{[A]}}''  }
    \label{fig:handpromptvqa_detail_nomask}
\end{figure}

\begin{figure}[tb!]
    \centering
    \includegraphics[width=0.98\columnwidth]{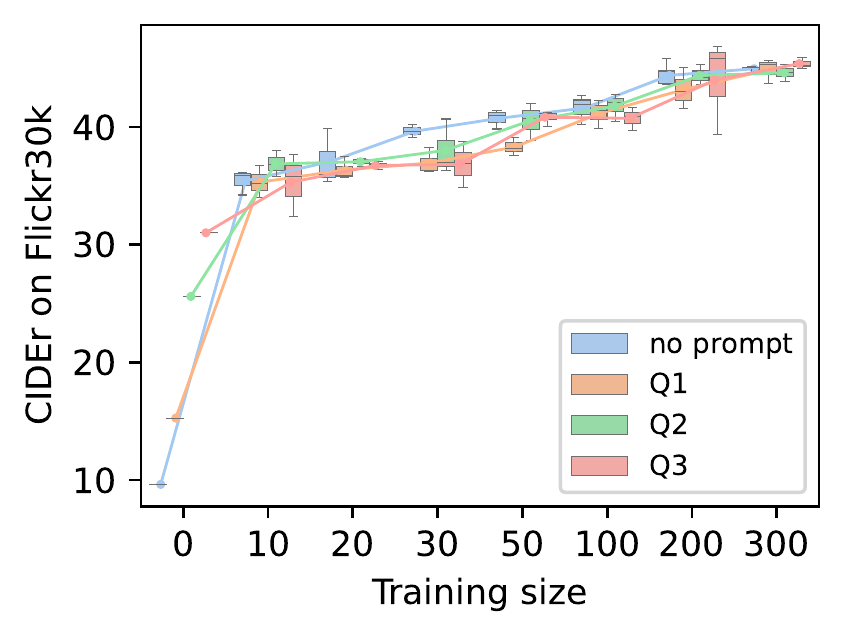}
    \caption{\textbf{Flickr30k results on hand-crafted prompts.}   
    }
    \label{fig:handpromptflickr_detail}
\end{figure}

\section{Prompt Study}
\label{app:prompt}

Tables~\ref{fig:handpromptvqa_detail}, \ref{fig:handpromptvqa_detail_nomask}, and \ref{fig:handpromptflickr_detail} show the results of each prompt on VQAv2 and Flickr30k with various training sizes.

\begin{table}[tb!]
\centering
\caption{\textbf{Few-shot results on different pre-training datasets.} We examine different pre-training datasets on each downstream tasks.
}
\resizebox{0.98\linewidth}{!}{
\begin{tabular}[t]{L{3.3cm}|C{2cm}C{2cm}C{2cm}}
\toprule
    \textbf{Dataset}  &\textbf{VQAv2}     &\textbf{GQA}   &\textbf{Flickr30k}\\
\midrule
MS COCO, VG &48.2   &32.2     &32.6      \\
Conceptual Captions &36.7   &25.9    &22.3   \\
\bottomrule
\end{tabular}}
\label{tab:cc}
\end{table}

\section{Effect of Pre-training Data}
We pre-train our model with different datasets: MS COCO and Visual Genome (VG), and Conceptual Captions (CC).
We investigate which pre-training dataset helps the downstream tasks in a few-shot manner.
In Table~\ref{tab:cc}, we observe that MS COCO and VG datasets are more helpful to the downstream tasks than CC.

\end{document}